\documentclass[10pt,twocolumn,letterpaper]{article}

\usepackage[pagenumbers]{wacv} 

\definecolor{wacvblue}{rgb}{0.21,0.49,0.74}
\usepackage[pagebackref,breaklinks,colorlinks,allcolors=wacvblue]{hyperref}
\usepackage[table]{xcolor} 
\usepackage{booktabs}
\usepackage{amsmath}
\usepackage{adjustbox}
\usepackage{array}

\newcount\heata     
\newcount\heatb     
\newcount\heatden   
\newcount\heattmp
\newcount\heatpct
\newcount\heatval

\definecolor{orange}{RGB}{255,180,129}
\definecolor{lightblue}{RGB}{200,235,255}

\makeatletter
\def\heatparsecount#1#2{%
  \edef\heatta{#2}%
  \expandafter\heatgetint\heatta.\@nil
  #1=\numexpr\heatint\relax
}
\def\heatgetint#1.#2\@nil{\def\heatint{#1}}
\makeatother

\newcommand{\setheatlimits}[2]{%
  \heatparsecount\heata{#1}%
  \heatparsecount\heatb{#2}%
  \def\heatas{#1}%
  \def\heatbs{#2}%
}

\makeatletter
\def\heatsetval#1{%
  \edef\heatta{#1}%
  \expandafter\heatgetint\heatta.\@nil
  \heatval=\numexpr\heatint\relax
}
\makeatother

\newcommand{\gsetheatlimits}[2]{%
  \heatparsecount\heattmp{#1}\global\heata=\heattmp
  \heatparsecount\heattmp{#2}\global\heatb=\heattmp
}
\newcommand{\rowheatlimits}[2]{\noalign{\gsetheatlimits{#1}{#2}}}

\newcolumntype{Y}[2]{>{\setheatlimits{#1}{#2}}r}

\newcommand{\heat}[1]{%
  \begingroup
  \heatsetval{#1}%
  \heatden=\numexpr\heatb-\heata\relax
  \ifnum\heatden=0
    \heatpct=50 
  \else
    \heattmp=\numexpr\heatval-\heata\relax
    \ifnum\heatden>0
      \ifnum\heattmp<0 \heattmp=0 \fi
      \ifnum\heattmp>\heatden \heattmp=\heatden \fi
    \else
      \ifnum\heattmp>0 \heattmp=0 \fi
      \ifnum\heattmp<\heatden \heattmp=\heatden \fi
    \fi
    \heatpct=\numexpr(100*\heattmp)/\heatden\relax 
    \ifnum\heatpct<0 \heatpct=0 \fi
    \ifnum\heatpct>100 \heatpct=100 \fi
  \fi
  \edef\colorspec{lightblue!\the\heatpct!orange}%
  \expandafter\cellcolor\expandafter{\colorspec}{#1}%
  \endgroup
}

\newcommand{\heatlog}[1]{%
  \begingroup
  \heatsetval{#1}%
  \heatden=\numexpr\heatb-\heata\relax
  \ifnum\heatden=0
    \heatpct=50
  \else
    \ifnum\heatden>0
      \ifnum\heatval<\heata \heattmp=\heata \else
      \ifnum\heatval>\heatb  \heattmp=\heatb  \else \heattmp=\heatval \fi \fi
    \else
      \ifnum\heatval>\heata \heattmp=\heata \else
      \ifnum\heatval<\heatb  \heattmp=\heatb  \else \heattmp=\heatval \fi \fi
    \fi
    \ifnum\heata>0 \ifnum\heatb>0 \ifnum\heattmp>0
      \edef\heatvc{\the\heattmp}%
      \edef\percstr{\fpeval{round(100*(ln(\heatvc)-ln(\heatas))/(ln(\heatbs)-ln(\heatas)),0)}}%
      \heatpct=\numexpr\percstr\relax
      \ifnum\heatpct<0 \heatpct=0 \fi
      \ifnum\heatpct>100 \heatpct=100 \fi
    \else\else\else
      \heattmp=\numexpr\heatval-\heata\relax
      \ifnum\heatden>0
        \ifnum\heattmp<0 \heattmp=0 \fi
        \ifnum\heattmp>\heatden \heattmp=\heatden \fi
      \else
        \ifnum\heattmp>0 \heattmp=0 \fi
        \ifnum\heattmp<\heatden \heattmp=\heatden \fi
      \fi
      \heatpct=\numexpr(100*\heattmp)/\heatden\relax
      \ifnum\heatpct<0 \heatpct=0 \fi
      \ifnum\heatpct>100 \heatpct=100 \fi
    \fi\fi\fi
  \fi
  \edef\colorspec{lightblue!\the\heatpct!orange}%
  \expandafter\cellcolor\expandafter{\colorspec}{#1}%
  \endgroup
}

\newcommand{\heatlogp}[1]{%
  \begingroup
  \heatsetval{#1}%
  \heatden=\numexpr\heatb-\heata\relax
  \ifnum\heatden=0
    \heatpct=50
  \else
    \ifnum\heatden>0
      \ifnum\heatval<\heata \heattmp=\heata \else
      \ifnum\heatval>\heatb  \heattmp=\heatb  \else \heattmp=\heatval \fi \fi
    \else
      \ifnum\heatval>\heata \heattmp=\heata \else
      \ifnum\heatval<\heatb  \heattmp=\heatb  \else \heattmp=\heatval \fi \fi
    \fi
    \ifnum\numexpr\heata+1\relax>0 \ifnum\numexpr\heatb+1\relax>0
      \edef\heatvc{\the\heattmp}%
      \edef\percstr{\fpeval{round(100*(ln(1+\heatvc)-ln(1+\heatas))/(ln(1+\heatbs)-ln(1+\heatas)),0)}}%
      \heatpct=\numexpr\percstr\relax
      \ifnum\heatpct<0 \heatpct=0 \fi
      \ifnum\heatpct>100 \heatpct=100 \fi
    \else\else
      \heattmp=\numexpr\heatval-\heata\relax
      \ifnum\heatden>0
        \ifnum\heattmp<0 \heattmp=0 \fi
        \ifnum\heattmp>\heatden \heattmp=\heatden \fi
      \else
        \ifnum\heattmp>0 \heattmp=0 \fi
        \ifnum\heattmp<\heatden \heattmp=\heatden \fi
      \fi
      \heatpct=\numexpr(100*\heattmp)/\heatden\relax
      \ifnum\heatpct<0 \heatpct=0 \fi
      \ifnum\heatpct>100 \heatpct=100 \fi
    \fi\fi
  \fi
  \edef\colorspec{lightblue!\the\heatpct!orange}%
  \expandafter\cellcolor\expandafter{\colorspec}{#1}%
  \endgroup
}

\setheatlimits{0}{100} 

\def\wacvPaperID{2693} 
\def\confName{WACV}
\def\confYear{2026}

\title{3D Gaussian Point Encoders}

\author{Jim James\\
Georgia Tech\\
{\tt\small jimjames@gatech.edu}
\and
Ben Wilson\\
Georgia Tech\\
\and
Simon Lucey\\
University of Adelaide\\
\and
James Hays\\
Georgia Tech\\
}

\begin{document}
\maketitle
\begin{abstract}
In this work, we introduce the 3D Gaussian Point Encoder, an explicit per-point embedding built on mixtures of learned 3D Gaussians. This \textit{explicit} geometric representation for 3D recognition tasks is a departure from widely used implicit representations such as PointNet. However, it is difficult to learn 3D Gaussian encoders in end-to-end fashion with standard optimizers. We develop optimization techniques based on \textit{natural gradients} and \textit{distillation} from PointNets to find a Gaussian Basis that can reconstruct PointNet activations. The resulting 3D Gaussian Point Encoders are faster and more parameter efficient than traditional PointNets. As in the 3D \textit{reconstruction} literature where there has been considerable interest in the move from implicit (e.g., NeRF) to explicit (e.g., Gaussian Splatting) representations, we can take advantage of computational geometry heuristics to accelerate 3D Gaussian Point Encoders further. We extend filtering techniques from 3D Gaussian Splatting to construct encoders that run \textbf{2.7$\times$} faster as a comparable accuracy PointNet while using \textbf{46\%} less memory and \textbf{88\%} fewer FLOPs. Furthermore, we demonstrate the effectiveness of 3D Gaussian Point Encoders as a component in Mamba3D, running \textbf{1.27$\times$} faster and achieving a reduction in memory and FLOPs by \textbf{42\%} and \textbf{54\%} respectively. 3D Gaussian Point Encoders are lightweight enough to achieve high framerates on CPU-only devices. Code is available at \url{https://github.com/jimtjames/3dGaussianPointEncoders}


\end{abstract}    
\section{Introduction}
\label{sec:intro}
Point cloud processing plays a crucial role in robotics and autonomous vehicles where LiDAR and related sensors capture three-dimensional spatial data. Since point clouds are unordered sets of points, deep networks designed for point cloud analysis must be permutation invariant to ensure consistent representations regardless of input ordering. Methods such as PointNet achieve this by employing symmetric aggregation functions, preserving the inherent structure of the data while enabling effective learning. This key feature has made PointNet ubiquitous in a variety of 3D tasks, including: classification \cite{qi2017pointnet++, ma2022rethinking}, detection \cite{lang2019pointpillars, zhou2018voxelnet, yan2018second}, and segmentation.


In PointNet, the majority of computational cost arises from per-point embedding, as it requires computing multiple large MLPs across a high number of points in each input point cloud. In contrast, the classifier stage applies MLPs only to a single global feature, making it relatively lightweight. To address this inefficiency, prior works have explored alternative approaches \cite{10550864}, such as LUTI-MLP \cite{sekikawa2019tabulated}, which replaces computationally expensive ReLU-MLP operations with lookup tables, and GPointNet \cite{gpointnet}, which employs single Gaussians. Although these methods greatly reduce FLOPs per sample compared to PointNet, their throughput on low-power platforms, such as CPU inference, remains limited. LUTI-MLP suffers from complex memory access patterns, while GPointNet requires evaluating a large number of Gaussian kernels, both of which hinder performance gains in resource-constrained environments.

Recently, explicit models based on mixtures of 3D Gaussians have gained traction in the view-synthesis literature due to their ability to efficiently represent volumetric data \cite{kerbl20233d, diolatzis2024n, fang2024mini, hanson2024speedy}. Several studies have leveraged the explicit nature of 3D Gaussians to reduce computational costs, employing techniques such as Gaussian pruning \cite{fang2024mini, niemeyer2024radsplat, hanson2024pup} and heuristic-based filtering of low-value Gaussian-point pairs \cite{ye20253d, hanson2024speedy}. These optimizations significantly accelerate inference compared to per-point coordinate networks.

In this work, we propose a novel \textbf{3D Gaussian Point Encoder}, a per-point embedding that integrates PointNet’s max-pooling aggregation with performance optimizations from view synthesis using mixtures of 3D Gaussians. By interpreting each dimension of PointNet’s embedding function as a volumetric representation, we leverage the capacity of 3D Gaussian mixtures to model volumes, enabling a lightweight approximation of a pre-trained PointNet. Moreover, we demonstrate it is possible to train this encoder end-to-end through Gaussian-specific natural gradient methods. Additionally, we exploit the explicit structure of Gaussians to enhance computational efficiency through filtering Gaussian-Point pairs. To assess the effectiveness of our approach, we conduct shape classification experiments on ModelNet40~\cite{wu20153d} and ScanObjectNN~\cite{uy-scanobjectnn-iccv19} while equipping our encoder with classical and modern classifiers from PointNet and Mamba3D.
In summary, our primary contributions are: 
\begin{enumerate}[(i)]
    \item We present a novel \textit{explicit} 3D representation as a drop in replacement for the \textit{implicit} PointNet representations which are ubiquitous in 3D scene understanding
    \item We discover that 3D Gaussian representations present significant optimization challenges when using off-the-shelf optimizers. We find two paths to overcome this roadblock -- distillation from PointNet teachers and direct optimization with Natural Gradients
    \item We show that explicit representations can benefit from geometric acceleration techniques, such as pairwise Gaussian-point filtering, inspired by the 3DGS literature
    \item We demonstrate that 3D Gaussian representations can achieve similar levels of accuracy to PointNet per-point embeddings, while achieving \textbf{2.7$\times$} higher throughput and \textbf{46\%} less memory. When integrated into Mamba3D, we achieve \textbf{1.27$\times$} the throughput and \textbf{42\%} less memory.
\end{enumerate}

\section{Related Work}
\label{sec:related_work}

\paragraph{Point Embeddings.} PointNet \cite{qi2017pointnet} is one of the first models to directly process point clouds, utilizing a per-point MLP with ReLU activations followed by a max-pooling operator. The output of this MLP serves as a spatial encoding for each point, while max-pooling aggregates these per-point embeddings into a single global feature representing the entire point cloud. Several works have extended PointNet to support hierarchical feature learning, including PointNet++ \cite{qi2017pointnet} and its modern variants \cite{ma2022rethinking, qian2022pointnext}. 


Several approaches rely on Transformers \cite{vaswani2017attention} as a component of their backbone, such as Point Cloud Transformer \cite{guo2021pct} and Point Transformer \cite{Zhao_2021_ICCV, wu2022point, wu2024point}. Transformer methods have the advantage of being possible to train from the vast quantity of unlabeled data via self-supervised learning, as done in Point-MAE \cite{pang2023masked} and Point-BERT \cite{yu2022point}. However, Transformers suffer from quadratic time complexity in sequence length, potentially resulting in inefficiency when processing large point sets. To resolve this issue, recent approaches instead utilize Mamba \cite{mamba, mamba2}, a structured state space model alternative to the Transformer with linear time complexity. PointMamba \cite{liang2025pointmamba} and PCM \cite{zhang2025point} aim to produce a vanilla Mamba-based model without a hierarchical encoder architecture. Most recently, Mamba3D \cite{han2024mamba3d} achieves near state-of-the-art performance on point classification, expanding upon PointMamba through the use of a bidirectional Mamba variant and local feature aggregation.

\paragraph{Efficient PointNet Variants.} A variety of prior works have explored efficient point encoders based on PointNet's per-point embedding with max-pooling framework. LUTI-MLP \cite{sekikawa2019tabulated} utilizes a lookup table per dimension of PointNet's embeddings formed followed by trilinear interpolation to form point embeddings. The lookup table is optimized during training time by discretizing and interpolating a pre-trained PointNet MLP, which is then voxelized at test time. This results in faster calculation of point embeddings compared to PointNet's MLP for 3D point clouds. However, as the input dimension increases, the runtime and memory cost grows exponentially due to the increased lookup table size and number of neighbors to interpolate. GPointNet \cite{gpointnet} instead represents each dimension of a point embedding via the likelihood of a single anisotropic Gaussian, resulting in an encoder requiring significantly fewer FLOPs per sample compared to PointNet.

\paragraph{Preconditioning and Natural Gradients.} Preconditioning is a technique in which an optimization problem is transformed to make it more amenable to numerical solvers. Several optimizers internally apply preconditioning to their gradients to stabilize training. These include the diagonal preconditioners in AdaGrad \cite{duchi2011adaptive} and Adam \cite{kingma2015adam}, as well as the block diagonal preconditioners in modern optimizers such as Shampoo \cite{gupta2018shampoo} and SOAP \cite{vyas2025soapimprovingstabilizingshampoo}.
One explicit form of preconditioning is Natural Gradients \cite{6790500}, a generalization of steepest descent for arbitrary metric spaces. This in contrast to standard gradient descent, where steps are considered with fixed Euclidean distance. \citet{6790500} demonstrated that given a metric, Natural Gradient descent can be viewed as preconditioning the gradients by the inverse of the metric space's Riemannian metric tensor. 

\paragraph{Mixtures of Gaussians as Approximators.} Methods utilizing mixtures of Gaussians, or more generally, radial basis functions \cite{park1993approximation}, have been widely studied. Classical works have used isotropic Gaussians to approximate volumes \cite{zhou2008real}. Most recently, a variety of works involving Gaussians have been applied to novel view synthesis. 3D Gaussian Splatting (3DGS) \cite{kerbl20233d} represents volumes via a mixture of anistropic Gaussians, and is able to render novel views signficantly faster than coordinate networks. The use of explicit Gaussians allows the method to exploit sparsity in real-world scenes. However, optimizing the set of Gaussians requires additional techniques compared to coordinate neural network-based approaches. \citet{niemeyer2024radsplat} notes that utilizing guidance from a pre-trained coordinate network can help train a more robust Gaussian representation to work around this issue.

Several works \cite{fang2024mini, hanson2024pup, fan2025lightgaussian, niemeyer2024radsplat} have reduced the computational costs of 3DGS via pruning and filtering. \citet{ye20253d} improve runtime by learning a truncation threshold on the Mahalanobis distance for each Gaussian, while \citet{hanson2024speedy} instead propose filtering before the Mahalanobis distance calculation by bounding each Gaussian with a rectangle or bounding via tiles, and then only computing points that fall within each bounding box or tile respectively.

\section{Method}
\label{sec:method}

We introduce the \textbf{3D Gaussian Point Encoder} (3DGPE), which replaces PointNet representations by simple, explicit 3D Gaussian functions for effective 3D shape classification. Surprisingly, we find that distilling point cloud features into a Gaussian-based network yields superior performance compared to directly optimizing Gaussian parameters. Additionally, our 3D Gaussian representation significantly reduces computational overhead by efficiently removing Gaussians and Gaussian-point pairs that do not meaningfully contribute to the final feature representation.


\begin{figure*}
    \centering
    \includegraphics[width=0.95\linewidth]{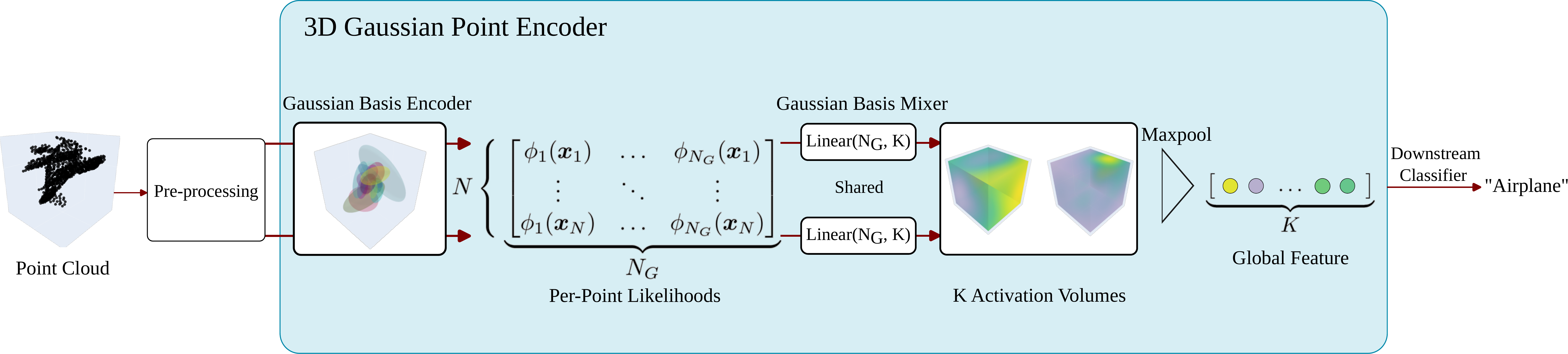}
    \caption{\textbf{Base architecture of 3DGPE.} An input point cloud is first pre-processed, such as by a T-Net or through Farthest Point Sampling and KNN. Afterwards, each input point is processed independently through the Gaussian Basis Encoder by first computing a set of Gaussian likelihoods, followed by the Gaussian Basis Mixer, mixing the likelihoods to form a set of embeddings for each activation volume. We max-pool across points to derive a global feature which is then passed to a downstream classifier, such as an MLP.}
    \label{fig:architecture}
\end{figure*}

Our 3D Gaussian Point Encoder comprises two key components: the \textbf{Gaussian Basis Encoder} and the \textbf{Gaussian Basis Mixer}. The Gaussian Basis Encoder encodes a point’s representation by computing its proximity to a set of 3D Gaussians, effectively capturing local geometric features. The Gaussian Basis Mixer then integrates these Gaussian-based features, transforming them into a richer and more expressive feature representation. This structured approach enables efficient and flexible encoding of spatial information for downstream tasks. In the following section, we outline their construction.

\subsection{Gaussian Basis Encoder}




The Gaussian Basis Encoder is a parametric function that maps input points from a 3D point cloud into a structured feature space using a set of learnable Gaussian functions. Given a point cloud \(\mathcal{X} = \{ x_i \}_{i=1}^{N}\), where each point \( x_i \in \mathbb{R}^3 \), the encoder represents the input as a mixture of spatial Gaussians. Each Gaussian component \( g \) is defined by a mean \( \boldsymbol{\mu}_g \in \mathbb{R}^3 \), which represents the center of the Gaussian in 3D space; a precision matrix (inverse of covariance matrix) \( \boldsymbol{\Sigma}_g^{-1} \in \mathbb{R}^{3 \times 3} \), modeling spatial extent; and a set of mixture coefficients \( \{\alpha_{g,k}\}_{k=1}^{K} \), where \( K \) denotes the number of activation volumes.

\paragraph{Covariance Parameterization.}
To ensure that \( \boldsymbol{\Sigma}_g^{-1} \) remains positive semi-definite, we parameterize it using the Cholesky decomposition \cite{diolatzis2024n}:
\begin{equation}
    \boldsymbol{\Sigma}_g^{-1} = \mathbf{L}_g \mathbf{L}_g^\top,
\end{equation}
where \( \mathbf{L}_g \) is a lower triangular matrix. This factorization guarantees valid covariance matrices while enabling efficient optimization. We parameterize the inverse directly to reduce the risk of numerical instability during training.

\paragraph{Feature Encoding.}
For each input point \( x \), we compute its unweighted Gaussian likelihood under each Gaussian \( g \) as follows:
\begin{equation}
    \phi_g(x) = \exp\left(-\frac{1}{2} (x - \boldsymbol{\mu}_g)^\top \boldsymbol{\Sigma}_g^{-1} (x - \boldsymbol{\mu}_g) \right).
\end{equation}
This function measures the proximity of \( x \) to the Gaussian distribution centered at \( \boldsymbol{\mu}_g \), with spatial spread determined by \( \boldsymbol{\Sigma}_g \).


\subsection{Gaussian Basis Mixer}
Following the Gaussian Basis Encoder, we introduce the Gaussian Basis Mixer, a critical component of our architecture that distinguishes it from prior methods such as GPointNet \cite{gpointnet}. Unlike previous approaches, the Gaussian Basis Mixer employs shared Gaussians across multiple activation volumes, effectively utilizing these Gaussians as basis functions. This design exploits redundancy, enhancing efficiency and enabling the network to represent complex activation volumes beyond simple ellipsoids.

Mathematically, the Gaussian Basis Mixer applies a linear layer using mixture coefficients to combine Gaussians and form activation volumes:
\begin{equation}
l_k(x) = \sum_{g=1}^{N_G} \alpha_{g,k} \phi_g(x) + b_k,
\end{equation}
where $b_k$ is a bias term unique to each activation volume. Following this, we maxpool across points to produce a permutation-invariant global feature.

Gaussian sharing significantly reduces latency and memory overhead as the input dimension increases, addressing a key computational bottleneck. The complexity of computing Gaussian likelihoods grows quadratically with input dimension due to the Mahalanobis distance computation. In contrast, the additional cost of uniquely recombining Gaussians for each activation volume scales only linearly with both the total number of Gaussians (\(N_G\)) and the number of activation volumes (\(K\)). This trade-off enables our architecture to efficiently handle high-dimensional inputs while maintaining expressiveness.

\begin{figure}
    \centering
    \includegraphics[width=0.95\linewidth]{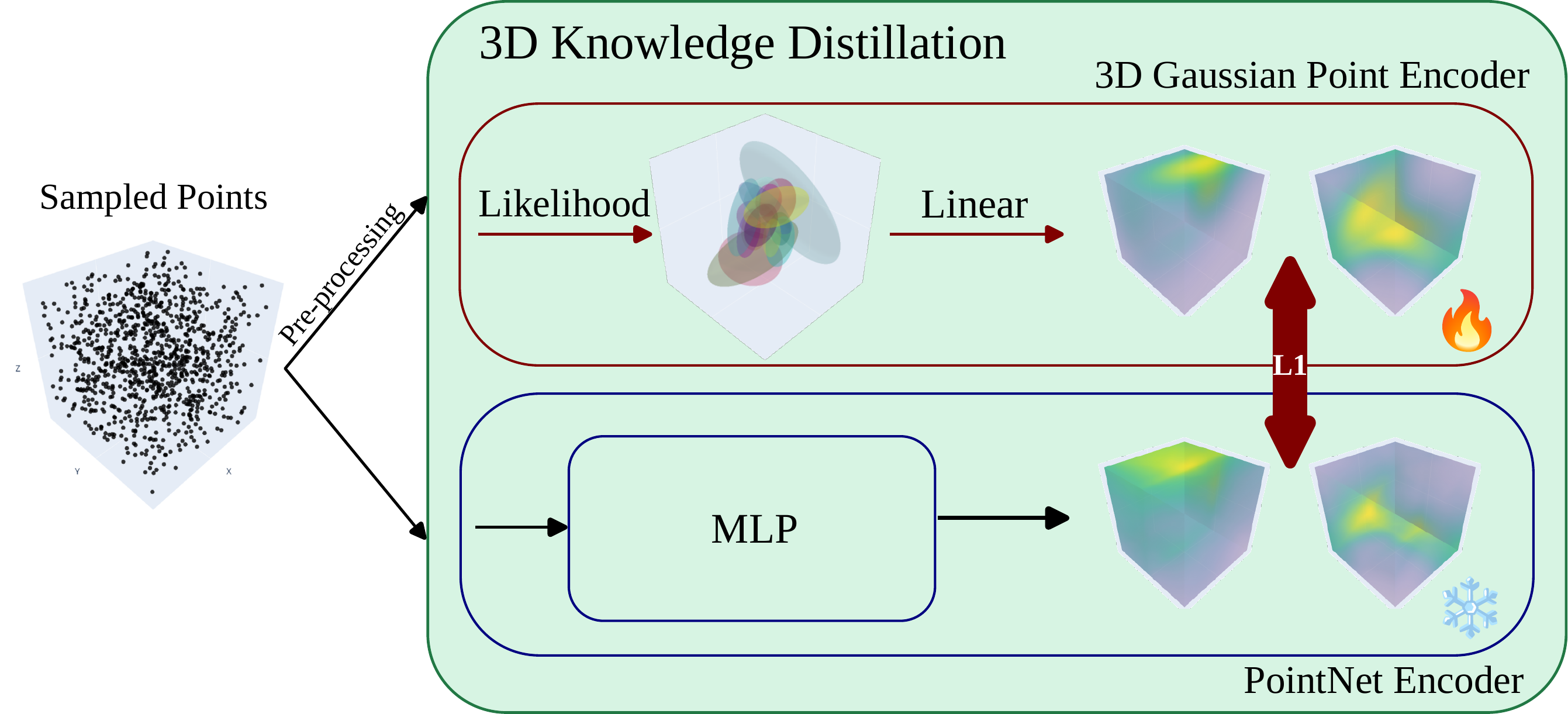}
    \caption{\textbf{Implicit to Explicit 3D Knowledge Distillation.} Points are sampled and pre-processed (T-Net or FPS + KNN) before being passed through each encoder. We then measure $L_1$ loss between the 3D Gaussian Point Encoder and PointNet per-point embeddings. Maroon outlines indicate trainable components, while blue indicates frozen components.}
    \label{fig:distillation}
\end{figure}

\subsection{Shape Classification Architecture}
We primarily experiment with utilizing our encoder with two classification architectures: PointNet \cite{qi2017pointnet} and Mamba3D \cite{han2024mamba3d}.
\paragraph{3D Gaussian Point Encoder with PointNet.}
The 3D Gaussian Point Encoder serves as the per-point embedding network; however, we add a few critical components to mimic a PointNet. The T-Net used in PointNet predicts a rotation matrix to achieve invariance to geometric transformations such as translation, rotation, and scaling. Since it's constructed from a PointNet, we are able to replace it with a 3DGPE network. We add the 3D Gaussian T-Net prior to passing the points through the backbone network. After then generating the global feature from the 3DGPE network, we compute our class logits by passing the global feature through a simple MLP classifier. This is equivalent in architecture to PointNet's classifier.



\paragraph{3D Gaussian Point Encoder with Mamba3D.}
Here, the 3D Gaussian Point Encoder serves as the patch encoder, generating feature embeddings for point sets formed through farthest point sampling and KNN-based grouping. After the point patches have been passed through the 3D Gaussian Point Encoder, we pass these embeddings through Mamba3D's middle encoder blocks while applying positional encodings. These blocks consist of a per-group normalization and feature aggregation operation, followed by a bi-directional state space model to capture global information about the point patch embeddings. We then compute class logits by applying an MLP classifier to the aggregated point embeddings. See \citet{han2024mamba3d} for more details.


%
\subsection{Optimization of 3D Gaussian Point Encoder}

We observe that end-to-end training of the 3D Gaussian Point Encoder with standard optimizers yields significantly lower performance compared to baseline models in PointNet and Mamba3D, as shown in \cref{tab:n_g-e2e}. We uncover two strategies to bypass this roadblock. The first is preconditioning the gradient via \textit{natural gradients}, and the second is to \textit{distill} the implicit geometry of PointNet features to the explicit geometry of 3DGPE.

\subsubsection{Natural Gradients for 3D Gaussians}

    Standard gradient descent minimizes loss by stepping in the direction of steepest decrease in the loss, assuming a fixed step size in Euclidean distance. Natural gradients \cite{6790500} generalize this by considering a different metric for step size, often resulting in faster convergence. \citet{6790500} notes that natural gradient descent can be performed via SGD while preconditioning the gradients by the inverse of the Riemannian metric tensor associated with a given parameter space, like so:

\begin{equation}
    \mathbf{x}^{t+1} = \mathbf{x}^{t} - \gamma \mathbf{G}^{-1} \nabla \mathcal{L}(\mathbf{x}^{t}),
\end{equation}
where $x^{t}$ is a parameter at iteration $t$, $\gamma$ is the learning rate, $\mathbf{G}$ is the Riemannian metric tensor, and $\mathcal{L}$ is the loss function. Gaussians as primitives admit two natural metrics.

\paragraph{Mahalanobis Distance.} Gaussian likelihoods are a function of the Mahalanobis distance of a query point to the mean with respect to the precision matrix. Accordingly, we can consider each Gaussian's mean an element in the metric space equipped with the Mahalanobis distance given its precision matrix. In this case, the Riemannian metric tensor is the precision matrix itself \cite{andyjonesnatgrad}. Thus, the natural gradient update for the $g$-th Gaussian's mean becomes:
\begin{equation}
   \boldsymbol{\mu}_g^{t+1} =  \boldsymbol{\mu}_g^{t} - \gamma \boldsymbol{\Sigma}_g^t \nabla \mathcal{L} \left(\boldsymbol{\mu}_g^{t}\right).
\end{equation}
See Sec. 1 of the supplemental material for an example.

\paragraph{Fisher Information Metric.} If we view each Gaussian primitive as a probability distribution, we can treat its mean and Cholesky decomposition parameters combined as an element in a parameter space defining probability distributions. One commonly used divergence for comparing such distributions is the KL divergence. The KL divergence can be approximated via a second order Taylor expansion to Fisher Information \cite{Tan_2025}. In this case, the Riemannian metric tensor is the Fisher Information Matrix $\boldsymbol{F}_g$, whose inverse can be easily computed in closed form and includes the same mean update as the Mahalanobis case \cite{Tan_2025}. Thus, the natural gradient update for the $g$-th Gaussian's parameters becomes:
\begin{equation}
   \left(\boldsymbol{\mu}_g^{t+1}, \boldsymbol{L}_g^{t+1}\right) =  \left( \boldsymbol{\mu}_g^{t}, \boldsymbol{L} _g^{t}\right) - \gamma \boldsymbol{F}^{-1}_g \nabla \mathcal{L}\left(\boldsymbol{\mu}_g^{t}, \boldsymbol{L}_g^{t}\right).
\end{equation}
In comparison to preconditioners applied by AdaGrad-inspired optimizers \cite{duchi2011adaptive, kingma2015adam}, neither of these preconditioning matrices are constrained to be diagonal, allowing them to capture more of the local geometry of each Gaussian. Furthermore, while optimizers like Shampoo \cite{gupta2018shampoo} and SOAP \cite{vyas2025soapimprovingstabilizingshampoo} instead utilize block diagonal preconditioning matrices, they recalculate the preconditioning matrices only for a subset of gradient updates for efficiency.

\subsubsection{Implicit to Explicit 3D Knowledge Distillation}





Our second approach is to first directly supervise our 3D Gaussian Point Encoder via a pre-trained PointNet-style per-point embedding. For our PointNet classification experiments, we perform this in three stages. Initially, we optimize the first Gaussian Basis Encoder, which serves as a T-Net, by sampling random points from the minimum bounding rectangular prism of the training set (\eg, the unit cube). We then minimize the $L_1$ loss between the per-point embeddings of PointNet's T-Net and those generated by the Gaussian Basis Encoder, prior to maxpooling.

Next, we enhance the 3D Gaussian Basis T-Net by incorporating a copy of the transform regressor MLP from the pre-trained T-Net. Once the per-point encodings for the T-Net are aligned, we proceed to optimize the main Gaussian Basis Encoder, which replaces the PointNet encoder. This optimization follows a similar process: sampling points from the bounding rectangular prism, computing the transformed points via each encoder’s T-Net, and minimizing the $L_1$ loss between the per-point embeddings produced by each encoder.

After this distillation phase, the model is trained end-to-end on the training set, utilizing a copy of the parent model's classifier. To preserve the learned per-point embeddings, we apply a reduced learning rate to both the T-Net encoder and the main encoder parameters, preventing significant deviations in their representations.

For our Mamba3D classification experiments, we only need two stages, as there is only one per-point embedding to distill. We optimize our 3DGPE network, which serves as a patch encoder, by instead sampling point clouds from the training dataset, and applying farthest point sampling and KNN-based grouping to generate in-distribution point patches. Similar to the PointNet case, we aim to minimize $L_1$ loss of the encoders' per-point embedding, following this up with end-to-end training with a copy of the parent model's middle encoder and classifier.

\subsection{Filtering via Explicit 3D Geometry}

Inspired by the pruning and filtering techniques in 3DGS \cite{fan2025lightgaussian, fang2024mini, diolatzis2024n, niemeyer2024radsplat}, we introduce Pairwise Gaussian-Point Filtering at inference-time in our encoder to further improve computational efficiency.

\begin{figure}
  \centering
  \begin{subfigure}[t]{0.3\linewidth}
      \includegraphics[width=\linewidth]{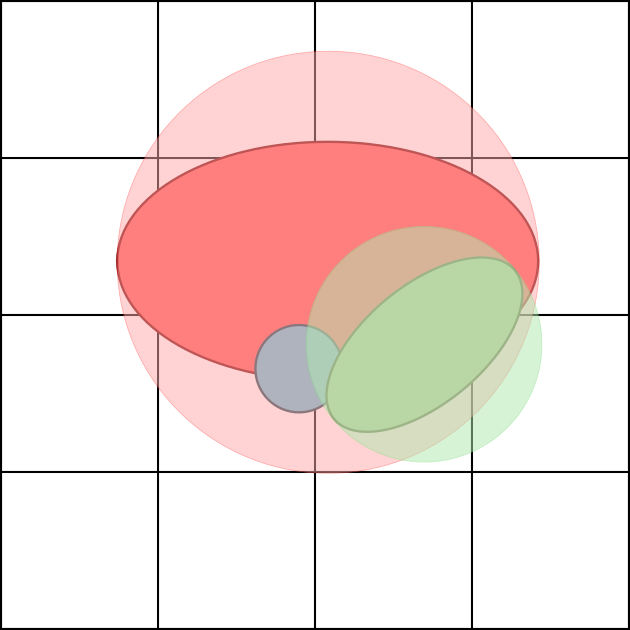}
    
    \caption{Distance Filtering}
    \label{fig:distance-filter}
  \end{subfigure}
  \hfill
  \begin{subfigure}[t]{0.3\linewidth}
      \includegraphics[width=\linewidth]{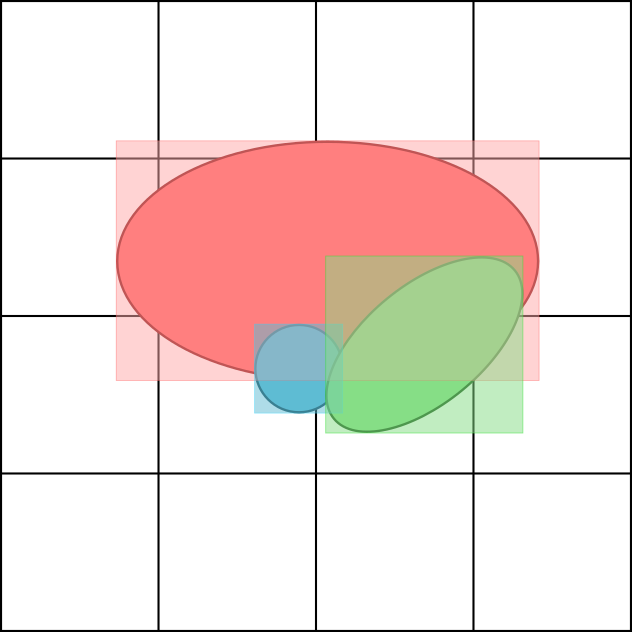}
    \caption{Bounding-box Filtering}
    \label{fig:bbox-filter}
  \end{subfigure}
  \hfill
  \begin{subfigure}[t]{0.3\linewidth}
      \includegraphics[width=\linewidth]{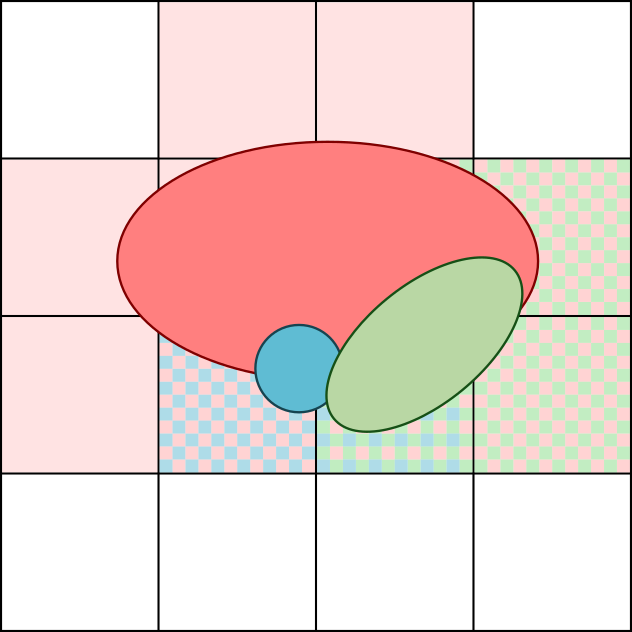}
    \caption{Voxel Filtering}
    \label{fig:bbox-filter}
  \end{subfigure}
  \caption{\textbf{Pairwise Gaussian-Point Filtering.} (a) Distance filtering only evaluates Gaussian-Point pairs within a radius of a Gaussian's mean. (b) Bounding-box Filtering evaluates Gaussian-Point pairs when a point falls within the axis-aligned bounding box center on a Gaussian. (c) Voxel Filtering evaluates Gaussian-Point pairs when a point lies in a voxel occupied with sufficiently high likelihood by a given Gaussian.}
  \label{fig:filtering}
\end{figure}


Computing the Mahalanobis distance has quadratic complexity in input dimension, making it relatively expensive. However, our experiments also reveal that a sizable percentage of the calculated Gaussian likelihoods are very small (see Fig. 2 in the supplemental). If the likelihood is sufficiently small, we can potentially filter it out and instead assume it to be zero. This requires a heuristic that is significantly faster to compute than Mahalanobis distance. We experiment with three heuristics:
\begin{enumerate}[(i)]
    \item \label{distance-based}\textbf{Distance Filtering.} We compute the Euclidean distance to each Gaussian's mean, which only requires linear complexity in dimension as opposed to quadratic.  We then threshold the distances by $2 \lambda_g \log\left( \frac{\alpha_{g, \text{max}}}{t_{\text{distance}}} \right)$, where $\lambda_g$ is the largest eigenvalue of the covariance matrix. We only evaluate the likelihood for Gaussian-point pairs below this distance. In essence, this method bounds an anistropic Gaussian with an isotropic Gaussian. This is variant of the method used by 3DGS \cite{kerbl20233d}.
    \item \label{bbox-based}\textbf{Bounding-box Filtering.} We compute the minimal axis-aligned bounding box for each Gaussian confidence ellipsoid given a threshold $t_{\text{bbox}}$, which requires bilinear computational cost in input dimension and number of Gaussians. Then, we check to see if a point falls within a bounding box before computing its likelihood. This is closely related to the ``SnugBox'' technique proposed by \citet{hanson2024speedy} for 3DGS, except extended to arbitrary dimensions rather than 2D.
    \item \label{voxel-based}\textbf{Voxel Filtering.} We coarsely voxelize the input volume by $D_{\text{voxel}}$ in each dimension and pre-compute the maximum weighted likelihood of each Gaussian for points falling within each voxel. We cache the list of Gaussians with weighted likelihood above a threshold $t_{\text{voxel}}$, and at runtime we only compute the likelihood of Gaussian-point pairs for each point's voxel's Gaussian list. This is related to the ``AccuTile'' technique proposed by \citet{hanson2024speedy} for 3DGS, except we pre-compute the weighted likelihoods rather than derive them with an interative algorithm.
\end{enumerate}
Each of these methods comes with various advantages and disadvantages. Method \ref{distance-based} benefits from re-using the computation of the difference between the points and the means, but is a poor heuristic if the Gaussians are highly anistropic. Method \ref{bbox-based} requires extra computation to determine bounding box occupancy, but more tightly encloses highly anistropic Gaussians than \ref{distance-based}. Finally, method \ref{voxel-based} can be implemented with very low computational costs at runtime using by a lookup table for the Gaussian lists and is the most accurate heuristic given a large enough $D_\text{voxel}$, but similar to LUTI-MLP \cite{sekikawa2019tabulated}, has exponential memory requirements in input dimension.
\subsection{Implementation Details}
\label{subsec:implement_details}
We implement our encoder in PyTorch, using \cite{Pytorch_Pointnet_Pointnet2} as a reference PointNet implementation for distillation experiments. Since our 3D Gaussian Point Encoder does not employ a feature transform (only an input transform via T-Net), we modify the PointNet implementation to remove the feature transform. We utilize Mamba3D's official release for our Mamba3D experiments, including their released weights for distillation. 

When training both our PointNet and Mamba3D variants end to end with natural gradients, we utilize SGD for the Gaussian parameters with a learning rate of $0.005$ on the means and $0.005$ on the Cholesky decomposition parameters, while using AdamW for the rest of the network. During distillation experiments, we train all components of our models using the AdamW optimizer with a learning rate of \(1.6 \times 10^{-3}\) for the means, \(5 \times 10^{-4}\) for the diagonal Cholesky elements, and \(1 \times 10^{-4}\) for the lower triangular Cholesky elements, mixture coefficients, and biases. MLPs used for 3D Gaussian T-Nets and classifiers utilize a learning rate of \(1 \times 10^{-4}\). The learning rates for the encoder parameters are reduced by a factor of 100 when fine-tuning following the initial distillation.

At test time, we pre-compute the precision matrices to avoid unnecessarily recomputing them for every point cloud. Furthermore, after computing the transformation matrix from the 3D Gaussian T-Net in our PointNet experiments, we apply the inverse transform to the Gaussian parameters rather than apply the transform to the points themselves. This reduces computational costs as there are significantly fewer Gaussians than points per sample.


\section{Experiments}
\label{sec:experiments}

\subsection{Shape Classification with PointNet and Mamba3D}
We benchmark our encoder on shape classification using the ModelNet40 \cite{wu20153d} and ScanObjectNN \cite{uy-scanobjectnn-iccv19} datasets. ModelNet40 consists of 9,843 training and 2,468 testing meshes of axis-aligned CAD models across 40 classes. We utilize the hardest ``PB T50 RS'' variant of ScanObjectNN, consisting of 11,416 training and 2,882 testing real-world 3D scans across 15 object classes. For both datasets, we reserve 25\% of the training samples as validation data for our ablations and hyperparameter selection. Following common practice, we report both class-averaged accuracy (mAcc.) and overall accuracy (OA) as our metrics. Furthermore, we measure FLOPs using FVCore. To gauge performance on varying hardware platforms, we measure GPU and CPU latency using PyTorch's profiler. GPU latency is measured on a single RTX 4070 Mobile GPU with a point cloud size of 2048 points, while CPU latency is measured on a low power ARM CPU (Rockchip RK3588) for methods that do not require custom CUDA extensions.

\subsubsection{Shape Classification Baselines}

For our PointNet experiments, we primarily compare against a PointNet with both an input transform and feature transform, as well as GPointNet, LUTI-MLP, and a PointNet pruned according to \cite{biswas20243d}. For our Mamba3D experiments, we instead compare against other Transformer and Mamba based architectures. Additionally, we include hierarchical architectures in PointNet++ \cite{qi2017pointnet++}, PointMLP \cite{ma2022rethinking}, and PointNeXT \cite{qian2022pointnext}, as well as a near state-of-the-art method in DeLA \cite{chen2023decoupled} for reference. All methods are evaluated without voting or cross-modal pre-training. We utilize rotation around the vertical axis for ScanObjectNN, and scaling by $\pm$ 20\% and translation by Gaussian noise with a standard deviation of 0.01 for ModelNet40. For both PointNet and Mamba3D experiments, we set $N_G$ to 32 and we utilize the Mahalanobis distance natural gradient. Our filtered PointNet model utilizes bounding-box filtering at test time with $t_{\text{bbox}}$ of 0.10.

\begin{table*}[!ht]
  \centering
  \caption{\textbf{Shape Classification Results.} FLOPs and Latency are computed per sample on ScanObjectNN with 2048 input points. X indicates that the model cannot be run on CPU, N indicates end-to-end with natural gradients, D indicates distilled, F indicates filtered. * indicates weights are not publicly available, so we cannot directly compare memory and latency.}
    \begin{adjustbox}{max width=\textwidth}
    \begin{tabular}{l Y{84}{94} Y{84}{94} Y{61}{91} Y{61}{91} Y{40}{0} Y{30}{1} Y{12}{0} Y{400}{37} Y{4000}{573}}
    \toprule
    & \multicolumn{2}{c}{ModelNet40} & \multicolumn{2}{c}{ScanObjectNN}  & FLOPs & Params & GPU Latency & CPU Latency  & Memory
      \\
    \cmidrule(lr){2-3}
    \cmidrule(lr){4-5}
    Method &  mAcc. (\%) & OA (\%) & mAcc. (\%) & OA (\%)   & (G) & (M) & (ms) & (ms) & (MB)\\
    \midrule
    \multicolumn{10}{c}{PointNet-Like Architectures} \\
    \midrule
    PointNet \cite{qi2017pointnet}& \heat{86.1} & \heat{90.0} & \heat{65.2} & \heat{69.0} & \heatlogp{0.891} & \heatlog{3.47} & \heatlogp{1.00}& \heatlog{110.2} & \heatlog{1057}\\
    PointNet (no FT)  & \textbf{\heat{86.4}} & \heat{90.2} & \heat{65.3} & \heat{69.3} & \heatlogp{0.582}  & \heatlog{1.61}  & \heatlogp{0.62} & \heatlog{69.3} & \heatlog{1039}\\
    GPointNet \cite{gpointnet} & \heat{84.3} & \heat{89.2} & \heat{58.4} & \heat{61.5} & \heatlogp{0.052} & \heatlog{1.34} & \heatlogp{14.81} & \heatlog{396.7} & \heatlog{2747}\\
    LUTI-MLP \cite{sekikawa2019tabulated} & \heat{85.9} & \heat{88.0} & \heat{60.9} & \heat{63.4} & \textbf{\heatlogp{0.032}} & \textbf{\heatlog{1.03}} & \heatlogp{3.67} & \heatlog{258.9} & \heatlog{4839}\\
    Pruned PointNet* \cite{biswas20243d} & - & \heat{88.2} & - & \textbf{\heat{71.7}} & - & \heatlog{1.36} & - & - & - \\
    \midrule
    \textbf{3DGPE (N)}   & \textbf{\heat{86.4}} & \heat{90.1} & \heat{65.5} & \heat{69.0} & \heatlogp{0.068} & \heatlog{1.39} & \heatlogp{0.44} & \heatlog{45.7} & \textbf{\heatlog{573}}\\
    \textbf{PointNet $\to$ 3DGPE (D)}   & \heat{86.1} & \textbf{\heat{90.3}} & \heat{65.3} & \heat{69.1} & \heatlogp{0.068} & \heatlog{1.39} & \heatlogp{0.44} & \heatlog{45.7} & \textbf{\heatlog{573}}\\
    \textbf{PointNet $\to$ 3DGPE (D + F)}   & \heat{85.3} & \heat{89.8} & \textbf{\heat{65.8}} & \heat{69.2} & \heatlogp{0.064} & \heatlog{1.39} & \textbf{\heatlogp{0.36}} & \textbf{\heatlog{37.6}} & \heatlog{605} \\
    \midrule
    \multicolumn{10}{c}{Dedicated \& Hierarchical Architectures} \\
    \midrule
    PointNet++ \cite{qi2017pointnet++}   & \heat{91.8} & \heat{89.1} & \heat{76.0} & \heat{77.8} & \heatlogp{1.68} & \heatlog{1.5} & \heatlogp{5.9} & \heatlog{403.4} & \heatlog{1215}\\
    PointMLP \cite{ma2022rethinking}  & \heat{91.3} & \textbf{\heat{94.1}} & \heat{83.9} & \heat{85.4} & \heatlogp{31.4} & \heatlog{12.6} & \heatlogp{7.7} & X & \heatlog{1801}\\
    PointNeXT \cite{qian2022pointnext}  & \heat{90.8} & \heat{93.2} & \heat{85.8} & \heat{87.7} & \heatlogp{1.6} & \textbf{\heatlog{1.4}} & \heatlogp{1.8} & X & \heatlog{1257} \\
    DeLA \cite{chen2023decoupled} & \textbf{\heat{92.2}} & \heat{94.0} & \textbf{\heat{89.3}} & \textbf{\heat{90.4}} & \textbf{\heatlogp{1.5}} & \heatlog{5.3} & \textbf{\heatlogp{0.9}} & X & \textbf{\heatlog{1177}}\\
    SimpleView \cite{goyal2021revisiting}  & - & \heat{93.9} & - & \heat{80.5} & - & - & - & X & -\\
    
    \midrule
    \multicolumn{10}{c}{Transformer and Mamba Architectures} \\
    \midrule
    PCT \cite{guo2021pct} & - & \heat{93.2} & - & - & \heatlogp{2.3} & \textbf{\heatlog{2.9}} & \heatlogp{14.8} & X & \heatlog{6677} \\
    PCM \cite{zhang2025point} & \textbf{\heat{90.7}} & \heat{93.4} & \heat{86.6} & \heat{88.1} & \heatlogp{45.0} & \heatlog{34.2} & \heatlogp{31.4} & X & \heatlog{5533} \\
    PointMamba \cite{liang2025pointmamba} & - & \heat{92.4} & - & \heat{84.9} & \heatlogp{3.1} & \heatlog{12.3} & \heatlogp{8.3} & X & \heatlog{1510}\\
    Mamba3D \cite{han2024mamba3d} & \heat{89.7} & \heat{93.3} & \textbf{\heat{90.6}} & \textbf{\heat{91.6}} & \heatlogp{3.9} & \heatlog{16.9} & \heatlogp{10.4} & X & \heatlog{1413}\\
    \midrule
    \textbf{3DGPE + Mamba3D (N)} & \heat{89.9} & \textbf{\heat{93.6}} & \heat{86.4} & \heat{88.0} & \heatlogp{1.8} & \heatlog{16.5} & \heatlogp{8.2} & X & \textbf{\heatlog{817}} \\
    \textbf{3DGPE + Mamba3D (D)} & \heat{89.8} & \heat{93.5} & \heat{86.6} & \heat{88.5} & \heatlogp{1.8} & \heatlog{16.5} & \heatlogp{8.2} & X & \textbf{\heatlog{817}} \\
    \textbf{3DGPE + Mamba3D (D + F)} & \heat{89.6} & \heat{93.3} & \heat{86.0} & \heat{88.3} & \textbf{\heatlogp{1.8}} & \heatlog{16.5} & \textbf{\heatlogp{7.8}} & X & \heatlog{853} \\
    
    \bottomrule
  \end{tabular}
  \end{adjustbox}
  \label{tab:classification}
\end{table*}

\subsubsection{Comparison to PointNet-like Architectures}
All PointNet-style classifiers perform comparably on ModelNet40. However, both GPointNet and LUTI-MLP underperform PointNet on ScanObjectNN compared to PointNet by approximately 7.8 and 5.9 percentage points respectively. We hypothesize that GPointNet's relatively low performance arises from its inability to model complex activation volumes, potentially making it harder to deal with the large perturbations present in ScanObjectNN. LUTI-MLP's lower performance may be also be a result of its modified T-Net, as it uses a $\tanh$ activation to constrain point clouds to fit in the unit cube, potentially resulting in deformation that interferes with its interpolation. In comparison, our 3D Gaussian Point Encoder performs comparably to PointNet, achieving the 2nd highest accuracy. 

Overall, we find that the 3D Gaussian Point Encoder with PointNet achieves the lowest latency out of all the models tested, achieving approximately \textbf{2.7$\times$} the throughput of a standard PointNet on a mobile GPU and \textbf{2.9$\times$} on a lower power CPU. Interestingly, despite the fact that both GPointNet and LUTI-MLP have lower FLOPs counts, both methods have substantially higher latency. Our latency advantage over these methods also holds on CPU, where both GPointNet and LUTI-MLP become prohibitively expensive, with throughputs under 4 samples per second. In the case of LUTI-MLP, this may be a result of the indexing operations required for interpolating the lookup table only being efficient with custom CUDA kernels. 

\subsubsection{Comparison to Advanced Architectures}
Among the Transformer and Mamba architectures, all methods perform comparably on ModelNet40. On ScanObjectNN, Mamba3D performs the best, with our 3D Gaussian Point Encoder performing similarly to PCM. Nonetheless, in comparison to these architectures, our model achieves the lowest FLOPs, latency, and memory. In fact, our model achieves the second lowest memory usage across all model types, highlighting how impactful the encoder design can be towards total memory usage. Compared to Mamba3D, our encoder reduces FLOPs by approximately \textbf{54\%} and memory by \textbf{42\%}, while increasing throughput by \textbf{1.27$\times$}.




\subsection{Ablations}
We ablate the impact of training each of the Gaussian parameters as well as optimization methods. All ablations are carried out on ScanObjectNN with evaluation performed on the validation set. An additional ablation on filtering methods is included in Sec. 2 in the supplemental.
\begin{table}[ht]
  \centering
  \caption{\textbf{Comparisons on $N_G$ and Optimization Methods.} Results are class-averaged accuracies on the validation split of ScanObjectNN averaged over 10 trials, listed alongside standard deviation. X denotes incompatibility. Mahalanobis and Fisher refer to the Mahalanobis distance and Fisher information metric natural gradients respectively.}
  \begin{adjustbox}{max width=\columnwidth}
  \begin{tabular}{*{6}{l}}
    \toprule
    & \multicolumn{5}{c}{\text{mAcc. (\%)}} \\
    \cmidrule(lr){2-6}
    {$N_G$} &  {Distill} & {Mahalanobis} & {Fisher} & {Adam} & {SOAP} \\
    \midrule
    \multicolumn{6}{c}{3DGPE} \\
    \midrule
    \rowheatlimits{50}{84}
    16 & \heat{68.6} $\pm$ 1.0 & \heat{65.9} $\pm$ 5.2 & \heat{66.3} $\pm$ 4.1 & \heat{43.7} $\pm$ 28.1 & \heat{60.0} $\pm$ 6.8 \\
    24 & \heat{71.1} $\pm$ 4.4 & \heat{72.4} $\pm$ 3.6 & \heat{69.9} $\pm$ 5.1 & \heat{65.8} $\pm$ 4.3 & \heat{68.3} $\pm$ 5.3\\
    32 & \heat{81.6} $\pm$ 4.9 & \heat{78.2} $\pm$ 4.3 & \heat{77.4} $\pm$ 2.4 & \heat{52.4} $\pm$ 23.6 & \heat{72.9} $\pm$ 6.5 \\
    64 & \heat{82.8} $\pm$ 3.7 & \heat{81.3} $\pm$ 4.2 & \heat{79.5} $\pm$ 3.1 & \heat{78.4} $\pm$ 5.0 & \heat{77.4} $\pm$ 7.2\\
    \midrule
    \multicolumn{6}{c}{3DGPE + Mamba3D}  \\
    \midrule
    \rowheatlimits{76}{88}
    16 & \heat{84.3} $\pm$ 2.4 & \heat{85.1} $\pm$ 2.2 & \heat{82.6} $\pm$ 8.5 & \heat{76.7} $\pm$ 5.3 & X \\
    24 & \heat{85.8} $\pm$ 3.2 & \heat{85.7} $\pm$ 3.7 & \heat{83.8} $\pm$ 4.6 & \heat{77.5} $\pm$ 6.9 & X \\
    32 & \heat{87.6} $\pm$ 2.0 & \heat{86.9} $\pm$ 1.5 & \heat{87.5} $\pm$ 2.1 & \heat{78.6} $\pm$ 6.6 & X \\
    64 & \heat{88.2} $\pm$ 2.1 & \heat{87.6} $\pm$ 2.6 & \heat{87.2} $\pm$ 4.2 & \heat{81.4} $\pm$ 7.6 & X \\
    
    \bottomrule
  \end{tabular}
  \end{adjustbox}
  \label{tab:n_g-e2e}
\end{table}

    

    

\subsubsection{Varying the Number of Gaussians}
We report the validation accuracies as we tweak the number of Gaussians, $N_G$ in \cref{tab:n_g-e2e}. Intuitively, performance generally increases as $N_G$ increases, as the Gaussian mixtures are better able to approximate PointNet's activation volumes. However, the performance does not significantly improve when increasing $N_G$ from 32 to 64.

\subsubsection{Optimization Techniques}
To validate the impact of natural gradients and distillation, we train both of our 3D Gaussian Point Encoder-based models from scratch with Adam \cite{kingma2015adam}, and only our PointNet model with SOAP \cite{vyas2025soapimprovingstabilizingshampoo}, as Mamba3D immediately returns NaN loss with it. In \cref{tab:n_g-e2e} we demonstrate that training both models end-to-end with Adam necessitates a substantial increase in $N_G$ to achieve acceptable performance, consequently resulting in increased computational cost. Moreover, we observe that, for most values of $N_G$, models trained end-to-end with standard optimizers exhibit significantly higher variability, and even their best-performing trials consistently underperform compared to trials utilizing either PointNet guidance or natural gradients. We hypothesize that this elevated variance arises from both the limited number of tunable parameters, which makes the optimization process more fragile, and heightened sensitivity to parameter initialization, especially with respect to the Gaussian means. We believe the preconditioned mean updates from both natural gradient methods, and to a lesser extent, SOAP,  allow them to mitigate some of this sensitivity.

\begin{table}[h]
  \centering
  \caption{\textbf{Ablations on Trainable Gaussian Parameters.} Results are class-averaged accuracies on the validation split of ScanObjectNN, and are averaged over 5 training runs with distillation. Performance generally goes down as more parameters are fixed.}
  \begin{adjustbox}{max width=0.65\columnwidth}
\begin{tabular}{c *{3}{c} l}
    \toprule
    & \multicolumn{3}{c}{Trainable Parameters} & 
      \\
    \cmidrule(lr){2-4}
    $N_G$ &  Mean & L. Triang. & Diag. & mAcc. (\%)\\
    \midrule
    \rowheatlimits{61}{69}
    \multirow{4}{*}[-2pt]{16} & \ding{51 } & \ding{51 } & \ding{51 } & \heat{68.6} \\
     & \ding{55} & \ding{51 } & \ding{51 } &  \heat{64.3} (-4.3)\\
     & \ding{55} & \ding{55} & \ding{51 } &  \heat{59.9} (-8.7)\\
     & \ding{55} & \ding{55} & \ding{55} &  \heat{60.6} (-8.0)\\
     \midrule
     \rowheatlimits{62}{78}
    \multirow{4}{*}[-2pt]{32} & \ding{51 } & \ding{51 } & \ding{51 } & \heat{81.6} \\
     & \ding{55} & \ding{51 } & \ding{51 } & \heat{62.2} (-19.4)\\
     & \ding{55} & \ding{55} & \ding{51 } & \heat{66.0}  (-15.6)\\
     & \ding{55} & \ding{55} & \ding{55} & \heat{65.5} (-16.1)\\
     \midrule
     \rowheatlimits{68}{83}
    \multirow{4}{*}[-2pt]{64} & \ding{51 } & \ding{51 } & \ding{51 } & \heat{82.8} \\
     & \ding{55} & \ding{51 } & \ding{51 } &  \heat{75.4} (-7.4)\\
     & \ding{55} & \ding{55} & \ding{51 } &  \heat{72.2} (-10.6)\\
     & \ding{55} & \ding{55} & \ding{55} &  \heat{69.7} (-13.1)\\
    \bottomrule
  \end{tabular}
  \end{adjustbox}
  \label{tab:trainable_params}
\end{table}

\subsubsection{Trainable Parameters}
We experiment with fixing the means, lower triangular covariance entries, and diagonal covariance entries of each of the Gaussians in our PointNet experiments. Fixing the lower triangular elements to zero makes the Mahalanobis Distance calculation more efficient but constrains the Gaussians to be axis-aligned, while fixing all the covariance entries constrains all Gaussians to have identity covariances. The results of this experiment are shown in \cref{tab:trainable_params}. In general, we find that all three Gaussian parameters contribute strongly to the model performance, with an especially sharp reduction in performance with diagonal covariance.

\section{Discussion and Conclusion}
\label{sec:conclusion}
In this paper, we introduced the \textit{3D Gaussian Point Encoder}, a novel point embedding architecture inspired by the explicit geometry of 3D Gaussian Splatting. Our experiments demonstrate that, when trained via natural gradients or 3D Knowledge Distillation, the 3D Gaussian Point Encoder achieves performance comparable to PointNet while significantly surpassing it in computational efficiency, delivering \textbf{2.7$\times$} higher throughput while using \textbf{46\%} less memory. Furthermore, the encoder integrates well into modern architectures like Mamba3D, improving throughput by $\textbf{1.27$\times$}$ and reducing memory by \textbf{42\%}.

\subsection{Limitations}
Our 3D Gaussian Point Encoder requires more careful optimization techniques than PointNet, and will likely not be suitable in cases where PointNet embeddings do not perform adequately. On Mamba3D experiments, 3DGPE was unable to optimize to the full level of performance as the original model. Furthermore, we only focus on classification. Higher dimensional inputs, such as decorators used in semantic segmentation and detection, may present unexpected challenges in fitting the Gaussian representation.

\clearpage
{
    \small
    \bibliographystyle{ieeenat_fullname}
    \bibliography{main}
}

\end{document}